%% file: neurips_2026.tex
\documentclass{article}
\usepackage{multirow}
\usepackage{graphicx}
\usepackage{enumitem}
\usepackage{marvosym} 
\usepackage{eso-pic}  
\usepackage{xcolor}
\usepackage[hidelinks]{hyperref}
\usepackage{caption}
\definecolor{projectblue}{RGB}{0,92,175}

\PassOptionsToPackage{numbers, compress}{natbib}
\usepackage[preprint]{neurips_2026}

\usepackage[utf8]{inputenc} 
\usepackage[T1]{fontenc}    
\usepackage{hyperref}       
\usepackage{url}            
\usepackage{booktabs}       
\usepackage{amsmath}        
\usepackage{amsfonts}       
\usepackage{nicefrac}       
\usepackage{microtype}      
\usepackage{xcolor}         
\usepackage{colortbl}       
\usepackage{float} 
\usepackage{placeins}
\definecolor{rankone}{HTML}{31A354}   
\definecolor{ranktwo}{HTML}{74C476}   
\definecolor{rankthree}{HTML}{C7E9C0} 
\definecolor{oursb}{HTML}{D6E8FA}     
\definecolor{headerbg}{HTML}{ECECEC}  
\definecolor{oomred}{HTML}{F4D0CC}    

\newcommand{\best}[1]{\textbf{#1}}
\newcommand{\snd}[1]{\underline{#1}}
\newcommand{\trd}[1]{#1}
\newcommand{\oom}{\cellcolor{oomred!70}\textcolor{black!55}{\scriptsize OOM}}

\title{IGGT4D: Streaming 4D Instance-Grounded Geometry Transformer}

\author{%
  \bfseries
  Zhengyu Zou$^{1,\dagger}$,
  Hao Li$^{2}$,
  Kuixuan Jiao$^{1,\dagger}$,
  Liu Liu$^{1,\ddagger}$,
  Tingyang Xiao$^{1}$, \\
  \bfseries
  Xiaolin Zhou$^{1}$,
  Fangzhou Hong$^{2}$,
  Zhizhong Su$^{1}$,
  Dingwen Zhang\textsuperscript{3,\normalfont\Letter},
  Ziwei Liu$^{2}$ \\[0.35em]
  \normalfont\footnotesize
  $^{1}$Horizon Robotics
  \qquad
  $^{2}$S-Lab, Nanyang Technological University \\[-0.1em]
  \normalfont\footnotesize
  $^{3}$Institute of Artificial Intelligence,
  Hefei Comprehensive National Science Center \\[0.15em]
  \normalfont\footnotesize
  Project Page:
  \href{https://iggt4d.github.io/}{%
  \textcolor{projectblue}{\texttt{https://iggt4d.github.io}}%
}
}

\begin{document}

\maketitle
\vspace{-1em} 
\AddToShipoutPictureFG*{%
  \AtPageLowerLeft{%
    \put(\LenToUnit{1.50in},\LenToUnit{0.56in}){%
      \rule{2.0in}{0.4pt}%
    }%
    \put(\LenToUnit{1.50in},\LenToUnit{0.42in}){%
      \footnotesize
      $^{\dagger}$ Intern at Horizon Robotics
      \qquad
      $^{\ddagger}$ Project Leader
      \qquad
      \Letter\ Corresponding Author%
    }%
  }%
}

\input{sections/abstract}

\input{sections/introduction}

\input{sections/related_work}

\input{sections/method}

\input{sections/dataset}

\input{sections/experiments}

\input{sections/conclusion}







\newpage
\bibliographystyle{unsrt}
\bibliography{refs}
\appendix

\input{sections/appendix}


\end{document}

%% file: sections/abstract.tex
\vspace{-6mm}
\begin{figure}[h]
    
    \centering
    \includegraphics[width=\textwidth]{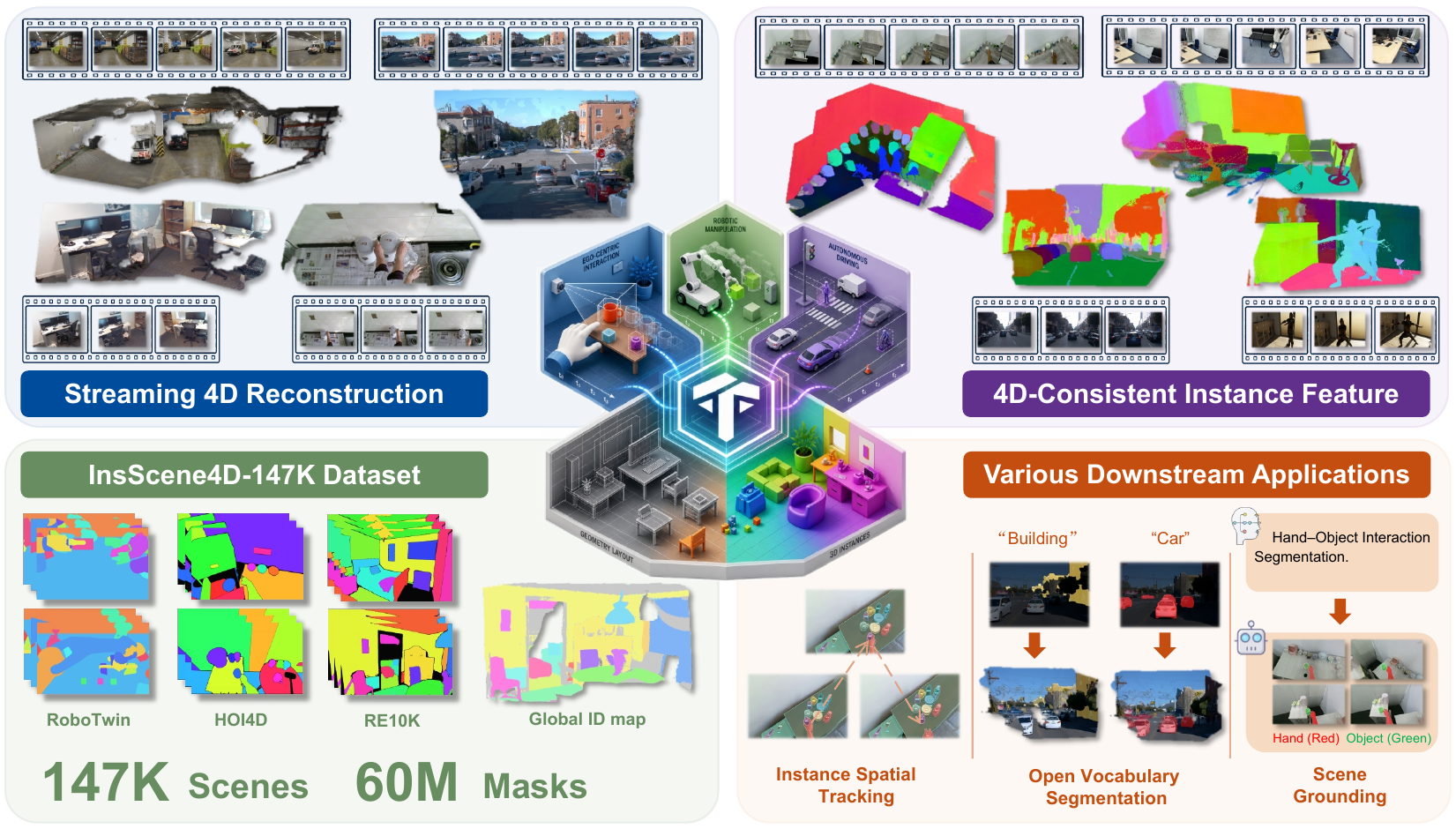}
    \caption{\textbf{Online Geometry-Instance Prediction with IGGT4D.}
    IGGT4D processes a dynamic video stream frame by frame to incrementally
    build a unified 4D representation that jointly captures camera motion,
    3D geometry, and temporally consistent instance features. This
    representation supports diverse downstream applications. We also
    construct \textbf{InsScene4D-147K}, a large-scale dataset with geometry-guided instance masks, to support training and evaluation.}
    \label{fig:teaser}
\end{figure}

\begin{abstract}
Real-world spatial intelligence requires agents to understand scenes from continuous video streams, where objects move, persist, disappear, and reappear over time. 
While recent spatial foundation models have enabled generalizable feed-forward 3D reconstruction, most streaming methods remain geometry-centric and lack temporally consistent object-level understanding. 
Meanwhile, existing semantic reconstruction and 3D-aware vision-language methods largely rely on externally extracted 2D semantic cues or loosely coupled geometry inputs, limiting unified geometry-instance learning in long dynamic scenes. 
In this paper, we propose \textbf{IGGT4D}, a streaming instance-grounded geometry Transformer for online 4D scene understanding. 
IGGT4D processes video frames sequentially, reuses historical context through causal spatial-temporal modeling, and incrementally updates a unified representation of camera motion, geometry, and object identity. 
This enables long-sequence feed-forward reconstruction with geometry-instance consistency in dynamic environments. 
To address the lack of high-quality 4D supervision, we further construct \textbf{InsScene4D-147K}, a large-scale dataset spanning real/synthetic and static/dynamic scenes, with RGB images, depth, poses, and temporally consistent instance masks generated by an automated geometry-guided annotation pipeline. 
Experiments on 3D reconstruction, pose estimation, instance spatial tracking, and open-vocabulary segmentation demonstrate that IGGT4D outperforms existing streaming baselines while maintaining scalable online inference for long dynamic sequences.
\end{abstract}

%% file: sections/introduction.tex
\section{Introduction}
\label{sec:intro}


Real-world spatial intelligence is inherently online and dynamic.
An embodied agent receives a continuous video stream in which objects move, become occluded, leave the field of view, and reappear over time.
Acting in such environments requires more than estimating camera motion and scene geometry: the agent must also maintain temporally consistent object identities.
We therefore formulate 4D scene understanding as streaming geometry-instance prediction, where scene geometry and object identities are updated jointly from long, dynamic video streams.

Spatial reconstruction and scene understanding are undergoing a shift from optimization-based pipelines to data-driven feed-forward prediction.
Traditional geometric pipelines, from offline SfM~\cite{schonberger2016structure,schonberger2016pixelwise} to online and neural SLAM~\cite{campos2021orb, geoflowslam, teed2021droid, zhu2022nice, irisslam}, rely on iterative per-scene optimization.
Similarly, semantic reconstruction systems, from offline scene graphs~\cite{werby2024hierarchical} to online semantic mapping~\cite{gu2024conceptgraphs}, often construct geometry and object-level cues through separate modules.
These systems are effective, but their scene-specific optimization limits long-sequence efficiency, while their modular design makes temporally consistent instance reasoning difficult in dynamic scenes.

Recent spatial foundation models~\cite{wang2025vggt,keetha2025mapanything,wang2025pi3,lin2025depth} address these limitations by learning generalizable priors for direct 3D prediction from images.
For online long-sequence perception, this paradigm is further moving from global fixed-set inference to streaming reconstruction~\cite{wang20253d,wang2025continuous,lan2025stream3r}.
However, streaming 3D prediction alone is insufficient for embodied agents.
Current streaming models remain largely geometry-centric: they estimate depth, pose, and point maps, but do not maintain temporally consistent object identities.
The missing capability is a feed-forward streaming model that treats object identity as a first-class prediction target together with geometry.


Why has streaming geometry-instance understanding remained underexplored? A key reason is the lack of suitable supervision.
2D vision-language models~\cite{radford2021learning,li2022language,ghiasi2022scaling,kirillov2023segment} provide scalable open-vocabulary cues, but their predictions are view-dependent and lack metric 3D grounding.
Semantic reconstruction methods~\cite{kerr2023lerf,peng2023openscene,takmaz2023openmask3d,sun2025uni3r,zhou2024feature} improve spatial coherence by lifting 2D semantics into 3D, but their semantic fidelity remains bounded by the capability of the external 2D predictors. 3D-aware vision-language models~\cite{hong20233d,zhu2024llava,fan2025vlm3rvisionlanguagemodelsaugmented,jiang2026spa3rpredictivespatialfield} incorporate geometry for spatial reasoning, but do not provide dense, temporally consistent geometry-instance supervision for training streaming models.
What is missing is large-scale 4D supervision that provides metric geometry, camera motion, and temporally consistent instance labels in a unified form.


To address these gaps, we propose \textbf{IGGT4D}, a streaming instance-grounded geometry Transformer for online 4D scene understanding. IGGT4D turns geometry-instance reconstruction from a full-sequence offline problem into a causal streaming prediction problem. It processes frames sequentially, reuses historical context through causal spatial-temporal modeling, and incrementally updates camera motion, scene geometry, and instance embeddings in a unified representation. By grounding instance association in reconstructed geometry, IGGT4D maintains object identities across viewpoint changes, occlusions, and reappearance.

We further construct \textbf{InsScene4D-147K}, a large-scale dataset for geometry-instance learning in 4D scenes. It spans real/synthetic and static/dynamic sources, and provides RGB images, depth maps, camera poses, point clouds, and sequence-level consistent instance masks. To reduce annotation cost, we design an automated geometry-guided annotation pipeline that produces multi-view consistent geometry and instance labels at scale. Our contributions are threefold:
\begin{itemize}[leftmargin=1.5em, itemsep=2pt, topsep=2pt]

\item \textbf{Streaming geometry-instance prediction.}
We formulate online 4D scene understanding as causal prediction of camera motion, scene geometry, and persistent object identities from continuous video streams.

\item \textbf{Object-consistent streaming reconstruction.}
We introduce IGGT4D, a feed-forward streaming Transformer that couples causal geometry modeling with geometry-grounded instance prediction and online clustering, maintaining consistent object identities without full-sequence inference.

\item \textbf{Scalable 4D instance supervision.}
We construct InsScene4D-147K, a large-scale real/synthetic and static/dynamic dataset with geometry-consistent instance annotations generated by a scalable reconstruction, projection, and mask-refinement pipeline.

\end{itemize}






%% file: sections/related_work.tex
\section{Related work}
\label{sec:related}

\paragraph{Streaming Spatial Foundation Models} 
Feed-forward spatial foundation models learn generalizable 3D priors for direct reconstruction from images~\cite{wang2025vggt,keetha2025mapanything,wang2025pi3,lin2025depth,wang2024dust3r,leroy2024grounding,peng2025omnivggt}. DUSt3R~\cite{wang2024dust3r} formulates dense 3D prediction as point-map regression~\cite{wang2024dust3r,leroy2024grounding}, while VGGT-style models extend this paradigm toward large-scale and modality-augmented visual geometry estimation~\cite{wang2025vggt,peng2025omnivggt}. However, these models are designed for fixed image sets; applying them to a growing video stream requires reprocessing the full history or using sliding windows, causing redundant computation and weakening long-range consistency. Recent streaming variants, including Spann3R~\cite{wang20253d}, MUSt3R~\cite{cabon2025must3r}, CUT3R~\cite{wang2025continuous}, and Stream3R~\cite{lan2025stream3r}, address this efficiency bottleneck by maintaining online scene state rather than rerunning global inference. While this reduces the overhead of incremental reconstruction, the maintained state remains optimized for geometric consistency rather than object persistence. Consequently, it lacks explicit object identities, instance masks, and cross-frame associations—elements indispensable for 4D scene understanding in long-duration dynamic videos.

\paragraph{Instance-Aware 3D Scene Understanding} Object-level scene understanding has also progressed from 2D open-vocabulary perception and per-scene semantic lifting toward unified 3D prediction. LangSplat~\cite{qin2024langsplat} and LangSurf~\cite{li2026langsurflanguageembeddedsurfacegaussians} attach language features to optimized 3D representations for open-vocabulary querying, while language-driven segmentation models such as LSeg~\cite{li2022language} predict semantic regions directly from images. These methods improve semantic accessibility, but they often depend on external 2D cues, per-scene optimization, or view-independent predictions, making it difficult to maintain coherent 3D object identities over time. More recent feed-forward scene understanding models aim to couple reconstruction with semantics or instances: LSM~\cite{fan2024large} predicts semantic radiance fields from unposed images, Uni3R~\cite{sun2025uni3r} predicts semantic 3D Gaussians from arbitrary multi-view inputs, and IGGT~\cite{li2025iggt} introduces instance-aware geometric prediction with 3D-consistent feature clustering. These models move closer to joint geometry-instance learning, but they are still built around fixed image sets and global cross-view attention. For long video streams, each update must attend to or recompute historical frames, making online inference expensive and preventing causal instance updates. IGGT4D targets the intersection of these two lines: it keeps the streaming efficiency of causal spatial foundation models while jointly predicting geometry and temporally consistent instance features for online 4D scene understanding.

%% file: sections/method.tex
\section{Method}
\label{sec:method}


We propose IGGT4D, a streaming instance-grounded geometry Transformer. Sec.~\ref{sec:problem_formulation} formalizes sequential feed-forward prediction. Sec.~\ref{sec:streaming_architecture} introduces the streaming architecture, and Sec.~\ref{sec:online_clustering} presents the efficient streaming clustering strategy. Sec.~\ref{sec:4d_scene_understanding} describes downstream 4D scene understanding applications, followed by the training objectives in Sec.~\ref{sec:training_objectives}.

\begin{figure}[t]
    \vspace{-6mm}
    \centering
    \includegraphics[width=0.98\textwidth]{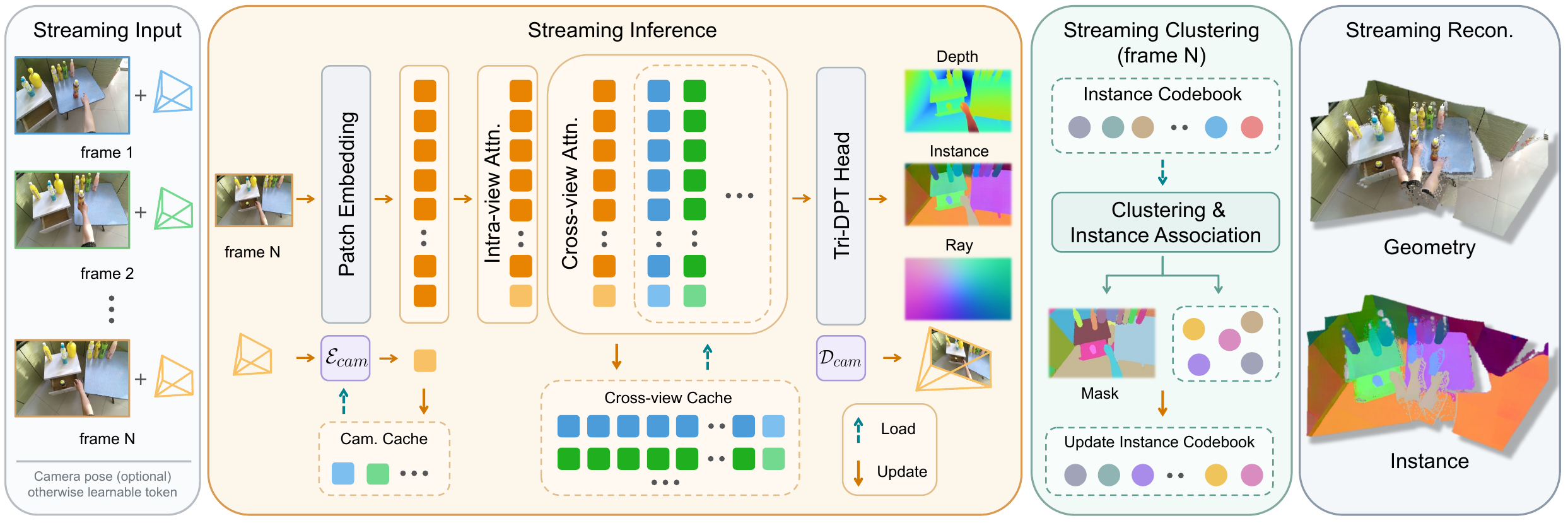}
    \caption{\textbf{Overview of IGGT4D.} Given a dynamic video sequence and optional camera poses, IGGT4D extracts a spatial-temporally consistent representation. A Tri-DPT head incrementally predicts geometry and instance features, and a streaming clustering algorithm derives instance masks for online 4D scene understanding.}
    \label{fig:method_architecture}
    \vspace{-6mm}
\end{figure}

\subsection{Problem Formulation and Notation}
\label{sec:problem_formulation}

Given an RGB image sequence $\{I_t\}_{t=1}^{N}$, where each image $I_t\in\mathbb{R}^{H\times W\times 3}$ and optional camera parameters $\{\tilde{\boldsymbol{\pi}}_t\}_{t=1}^{N}$ with $\tilde{\boldsymbol{\pi}}_t\in\mathbb{R}^{9}$, IGGT4D (denoted as $\mathcal{F}_{\theta}$) performs feed-forward sequential inference for streaming 4D geometry and instance reconstruction:
\begin{equation}
\mathcal{F}_{\theta}:
\left(I_t,\tilde{\boldsymbol{\pi}}_t\right)
\mapsto
\left(\boldsymbol{\pi}_t,R_t,D_t,S_t\right),
\quad t=1,\ldots,N.
\label{eq:iggt4d_io}
\end{equation}
For each frame, IGGT4D outputs camera parameters $\boldsymbol{\pi}_t=[\mathbf{t}_t,\mathbf{q}_t,\mathbf{f}_t]\in\mathbb{R}^{9}$, a ray map $R_t=[O_t,V_t]\in\mathbb{R}^{H_r\times W_r\times 6}$, a depth map $D_t\in\mathbb{R}^{H\times W}$, and an instance feature map $S_t\in\mathbb{R}^{H\times W\times 8}$. Here, $\mathbf{t}_t\in\mathbb{R}^{3}, \mathbf{q}_t\in\mathbb{R}^{4}, \mathbf{f}_t\in\mathbb{R}^{2}$ denote translation, rotation, and field-of-view, respectively, while $O_t\in\mathbb{R}^{H_r\times W_r\times 3}, V_t\in\mathbb{R}^{H_r\times W_r\times 3}$ represent ray origins and directions. As images are sequentially streamed in, frame-wise predictions are integrated into a spatial-temporal consistent scene representation, enabling online 4D reconstruction and understanding.
\subsection{Streaming Instance-Grounded Geometric Transformer}
\label{sec:streaming_architecture}

\textbf{Causal Geometry-Instance Transformer.}
We adapt the unified geometric Transformer of DA3~\cite{lin2025depth} from fixed-set 3D prediction to causal streaming 4D understanding.
Each image is encoded into image tokens and concatenated with a camera token, obtained from $\mathcal{E}_{\mathrm{cam}}$ when camera parameters are available or from a shared learnable token otherwise.
The resulting frame-level tokens are processed by 40 Transformer blocks with interleaved intra-view and cross-view attention, producing multi-scale features $\{\mathbf{F}_t^{(l)}\}_{l=1}^{4}$.
Unlike DA3's bidirectional fixed-view reasoning, IGGT4D imposes causal masks on cross-camera and cross-view attention, so each frame only attends to current and past observations.
The same constraint is used during training to simulate streaming inference.
At inference time, frames are processed sequentially, with camera and cross-view KV caches reusing historical context without redundant recomputation.
The resulting representation supports scalable long-sequence geometry and instance prediction.



\textbf{Tri-DPT Geometry-Instance Head.}
We design a Tri-DPT head to jointly decode depth $D_t$, ray map $R_t$, and instance feature map $S_t$ from the streaming multi-scale tokens $\{\mathbf{F}_t^{(l)}\}_{l=1}^{4}$.
All branches progressively recover spatial resolution, while the instance branch is coupled with geometric branches through geometry-aware attention.
By using depth and ray features as structural priors, the head grounds instance embeddings in 3D geometry and improves object identity consistency across time.


\subsection{Efficient Streaming Instance Clustering}
\label{sec:online_clustering}
Offline clustering, such as HDBSCAN~\cite{mcinnes2017hdbscan} used in
IGGT~\cite{li2025iggt}, incurs prohibitive quadratic complexity on long
sequences. To enable online inference, we introduce a two-stage
streaming clustering strategy that maintains a lightweight global
instance codebook
$\mathcal{C}_t=\{(\mathbf{c}_k,a_k)\}_{k=1}^{K_t}$, where
$\mathbf{c}_k$ is the feature center of instance $k$ and $a_k$ is its
accumulated pixel count.

\begin{figure}[t]
    \vspace{-6mm}
    \centering
    \includegraphics[width=0.98\textwidth]{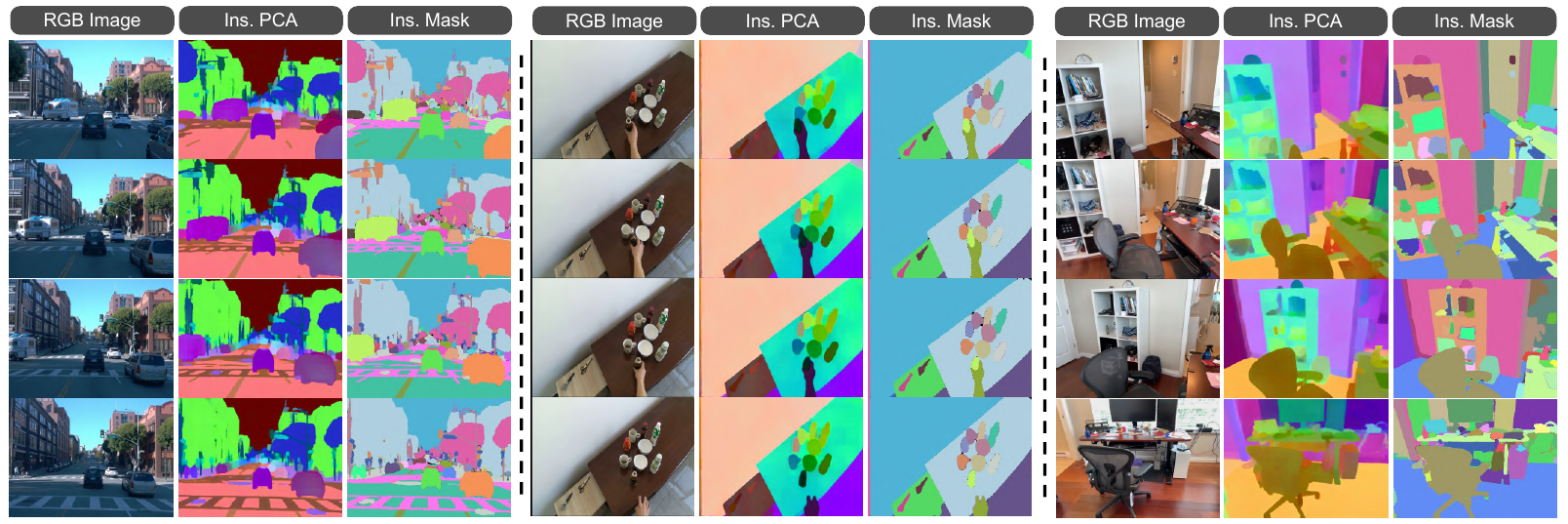}
    \caption{\textbf{Instance Feature and Mask Visualization.} We visualize the 3D-consistent instance feature PCA results alongside the corresponding instance masks generated by our streaming clustering.}
    \label{fig:pca_mask_vis}
    \vspace{-6mm}
\end{figure}

First, intra-frame clustering extracts local masks $M_{t,i}$ and centers $\mathbf{c}_{t,i}$. For an unassigned pixel $p$, we group pixels with high cosine similarity ($\alpha_q = \mathbf{s}_p^\top \mathbf{s}_q \geq \tau_s$) into a core region, then expand it via connected components over a looser threshold ($\tau_l$) to form $M_{t,i}$. Next, we match these local centers with the global codebook. Unmatched instances become new global entries, while unmatched background regions are naturally ignored as distinct static entities, implicitly decoupling dynamic foreground from the static background. For matches, $M_{t,i}$ inherits the global ID $k$ as a 4D-consistent mask $M_{t,k}$, and its center is updated via area-weighted fusion:
\begin{equation}
\mathbf{c}_k \leftarrow
\mathrm{norm}\left( \frac{a_k\mathbf{c}_k+|M_{t,k}|\mathbf{c}_{t,i}}{a_k+|M_{t,k}|} \right),
\quad
a_k\leftarrow a_k+|M_{t,k}|.
\label{eq:clustering_update}
\end{equation}
This constant-time center update translates 4D-consistent instance features into high-quality tracking masks (Fig.~\ref{fig:pca_mask_vis}), providing a lightweight mechanism for maintaining instance consistency under object motion and partial occlusion, without explicit motion modeling.

\subsection{4D Scene Understanding}
\label{sec:4d_scene_understanding}


The 4D-consistent masks $M_{t,k}$ and temporally linked instance features $\mathbf{c}_k$ provide a reusable object-level representation for downstream scene understanding tasks. First, the persistent instance correspondences across frames enable stable spatial tracking without additional association heuristics. Second, by using $M_{t,k}$ to aggregate per-frame 2D vision-language features~\cite{radford2021learning, ghiasi2022scaling}, we obtain spatially and temporally consistent language features $\mathbf{f}_{t,k}^{\mathrm{lang}}$ for open-vocabulary semantic segmentation. Finally, these 4D-consistent masks and features can be provided to Large Multimodal Models (LMMs)~\cite{yang2025qwen3, bai2025qwen3} to support 4D scene grounding, including tasks such as dynamic object segmentation and out-of-view object reasoning. Additional implementation details and qualitative results are provided in the supplementary material.


\subsection{Training Objectives}
\label{sec:training_objectives}

\textbf{First-frame Geometric Normalization.} In streaming reconstruction, estimating scale from the entire sequence, as in offline settings, can introduce scale ambiguity. To avoid this ambiguity and promote spatial-temporal consistency, we apply first-frame-based geometric normalization during training: all poses are aligned to the first-frame camera coordinate system, and the first-frame point cloud $\mathcal{P}_1$ is used to compute a sequence-level scale $s=\frac{1}{|\mathcal{P}_1|}\sum_{\mathbf{p}\in\mathcal{P}_1}\|\mathbf{p}\|_2$, which is then applied to all geometric ground truth across the sequence.

\textbf{Geometry Objectives.} We supervise geometry with depth, ray, point, and camera losses: $\mathcal{L}_{\mathrm{geo}} = \mathcal{L}_D + \mathcal{L}_R + \mathcal{L}_P + \mathcal{L}_{\pi}$. The depth loss $\mathcal{L}_D$ combines L1 regression, confidence $C_t$ regularization, and spatial gradient $\mathcal{L}_{\mathrm{grad}}$ consistency over valid pixels $V$:
\begin{equation}
\mathcal{L}_D = \frac{1}{|V|} \sum_{p \in V} \left( |D_t(p) - D_t^\ast(p)| \big( 1 + \lambda_{\mathrm{conf}} C_t(p) \big) - \lambda_{\mathrm{conf}} \alpha \log C_t(p) \right) + \lambda_{\mathrm{grad}} \mathcal{L}_{\mathrm{grad}},
\end{equation}
where $\mathcal{L}_{\mathrm{grad}} = \|\nabla D_t - \nabla D_t^\ast\|_1$ is an L1 gradient loss over spatial neighbors, $V$ is the set of valid pixels, and $\alpha$ is a scaling factor. For the ray map $R_t=[O_t,V_t]$, we apply L1 loss to origins and directions: $\mathcal{L}_R = \|O_t-O_t^\ast\|_1 + \|V_t-V_t^\ast\|_1$. We further couple depth and rays by reconstructing 3D points $P_t=O_t+D_tV_t$ and minimizing $\mathcal{L}_P=\|P_t-P_t^\ast\|_1$. The camera loss is $\mathcal{L}_{\pi}=\|\boldsymbol{\pi}_t-\boldsymbol{\pi}_t^\ast\|_1$.

\textbf{Instance Objective.} We optimize the L2-normalized instance features using a multi-view contrastive loss. For each annotated mask, we define its prototype $\mu$ as the mean pixel feature. The objective enforces intra-view ($u=v$) constraints to group pixels $\mathbf{f}_p$ toward their instance center $\mu_k^v$ and repel different centers. It also applies cross-view ($u \neq v$) constraints to pull matching centers across frames and push different centers apart:
\begin{equation}
\begin{aligned}
\mathcal{L}_{\mathrm{ins}} &= \sum_{v} \Big( \lambda_{\mathrm{pull}}^{\mathrm{in}} \sum_{p \in M_k^v} \big[ \|\mathbf{f}_p - \mu_k^v\|_2 - \delta_{\mathrm{pull}}^{\mathrm{in}} \big]_+ + \lambda_{\mathrm{push}}^{\mathrm{in}} \sum_{k \neq j} \big[ \delta_{\mathrm{push}}^{\mathrm{in}} - \|\mu_k^v - \mu_j^v\|_2 \big]_+ \Big) \\
&+ \sum_{u \neq v} \Big( \lambda_{\mathrm{pull}}^{\mathrm{cr}} \sum_{k} \big[ \|\mu_k^u - \mu_k^v\|_2 - \delta_{\mathrm{pull}}^{\mathrm{cr}} \big]_+ + \lambda_{\mathrm{push}}^{\mathrm{cr}} \sum_{k \neq j} \big[ \delta_{\mathrm{push}}^{\mathrm{cr}} - \|\mu_k^u - \mu_j^v\|_2 \big]_+ \Big),
\end{aligned}
\end{equation}
where $[ \cdot ]_+ = \max(0, \cdot)$, and $\delta$ are their respective margins. The final objective is $\mathcal{L}=\mathcal{L}_{\mathrm{geo}}+\lambda_{\mathrm{ins}}\mathcal{L}_{\mathrm{ins}}$.


%% file: sections/dataset.tex
\section{InsScene4D-147K Dataset}
\label{sec:instance_annotation}

\begin{figure}[t]
    \vspace{-6mm}
    \centering
    \includegraphics[width=0.96\textwidth]{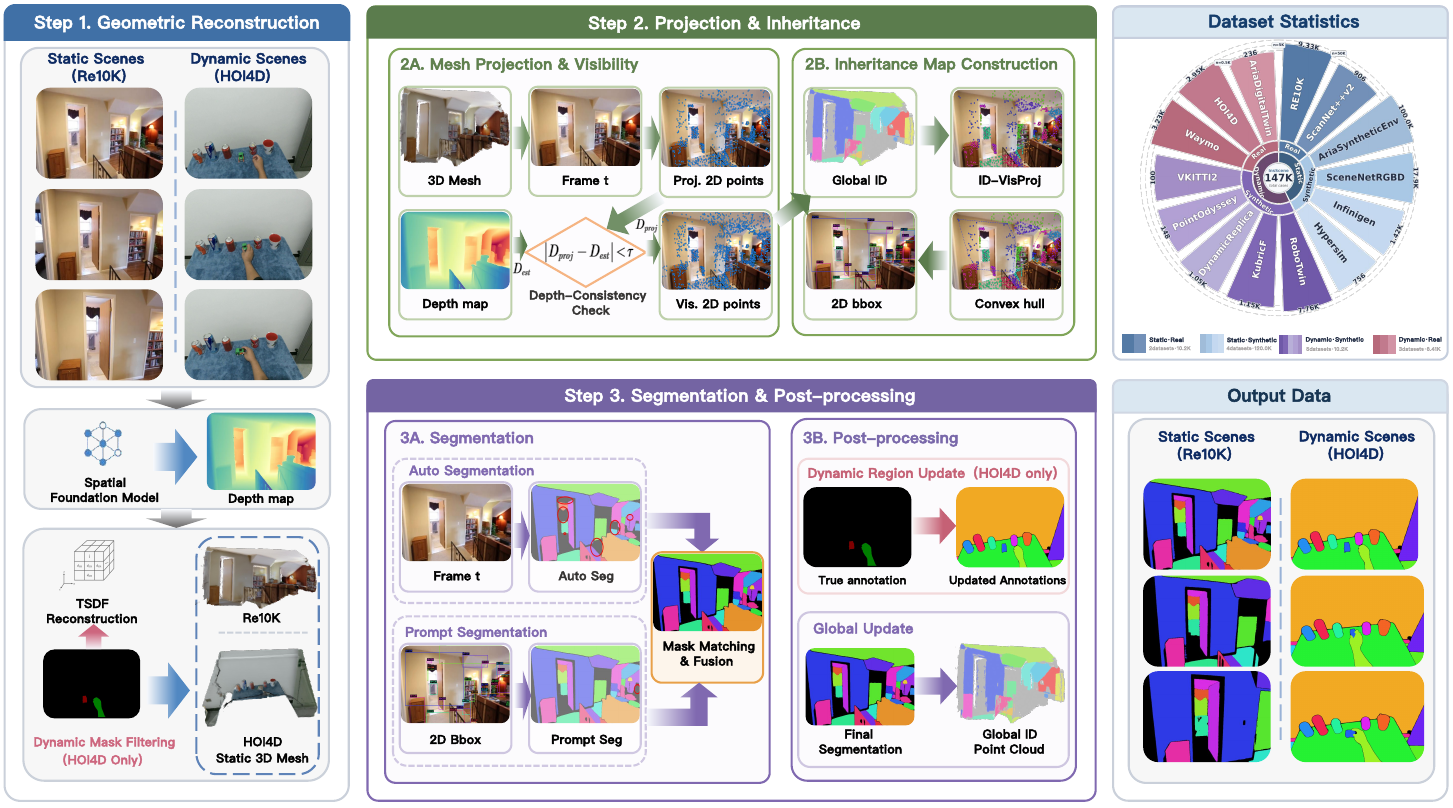}
    \caption{\textbf{InsScene4D-147K data curation pipeline.} Real/synthetic and static/dynamic sources are processed through 3D reconstruction, projection-based ID inheritance, and segmentation-based global ID refinement. The right panel summarizes the dataset splits.}
    \label{fig:data_pipeline}
    \vspace{-0.25in}
\end{figure}


We construct \textbf{InsScene4D-147K}, a large-scale dataset for online 4D scene understanding, comprising \emph{147K} curated video sequences. The dataset spans four domains: static-real, static-synthetic, dynamic-real, and dynamic-synthetic. Specifically, the static-real split includes RealEstate10K~\cite{zhou2018stereo} and ScanNet++~\cite{yeshwanth2023scannet++}; the static-synthetic split includes Aria Synthetic Environments~\cite{pan2023aria}, SceneNet~\cite{mccormac2017scenenet}, Infinigen~\cite{raistrick2024infinigen}, and Hypersim~\cite{roberts2021hypersim}; the dynamic-real split includes HOI4D~\cite{liu2022hoi4d}, Waymo~\cite{mei2022waymo}, and Aria Digital Twin~\cite{pan2023aria}; and the dynamic-synthetic split includes RoboTwin~2.0~\cite{chen2025robotwin}, Kubric~\cite{greff2022kubric}, Dynamic Replica~\cite{karaev2023dynamicstereo}, PointOdyssey~\cite{zheng2023pointodyssey}, and VKITTI2~\cite{cabon2020virtual}. Each sequence provides RGB images, depth maps, camera poses, point clouds, and 4D-consistent instance masks with persistent object identities across frames. Fig.~\ref{fig:data_pipeline} illustrates our geometry-guided annotation pipeline for producing temporally consistent instance supervision at scale.


For static captures, we design a curation pipeline that turns offline geometry into temporally consistent instance supervision. 
We first estimate multi-view consistent depth with DA3~\cite{lin2025depth}, conditioned on offline ground-truth camera poses to avoid long-sequence pose drift. 
A static 3D mesh is then reconstructed via TSDF fusion~\cite{curless1996volumetric}, which aggregates depth predictions across views to reduce estimation variance and suppress outlier noise. 
This process yields high-quality, multi-view consistent geometric pseudo-labels for subsequent instance annotation.


For each frame, we project mesh vertices onto the image plane, retain visible vertices by depth consistency, and reverse-map their global IDs to form an \emph{inheritance map}. SAM2~\cite{ravi2024sam} provides category-agnostic masks, which are filtered, de-overlapped, and matched to inherited regions by IoU. Confident matches inherit existing IDs, ambiguous matches use the best-overlap ID, and unmatched masks are initialized as new instances. To prevent stale ID propagation, an object is marked as disappeared if its projected area drops sharply for $N{=}5$ consecutive frames.

For dynamic scenes such as HOI4D, we estimate depth and camera poses with DA3, remove dynamic regions using the provided masks, and reconstruct the static 3D mesh from the remaining regions. 
We then apply the same annotation pipeline, while using available dynamic-object annotations to override projected pseudo-labels in dynamic regions. 
For simulation data, the source datasets provide RGB-D sequences with instance-consistent masks, which we incorporate into the synthetic split.

%% file: sections/experiments.tex
\section{Experiments}



We evaluate our method across multiple tasks against a broad range of state-of-the-art baselines. All experiments are conducted on an NVIDIA RTX 5090 GPU with 32 GB of memory.

\subsection{Evaluation of Camera Pose Estimation and 3D Reconstruction}

Following the evaluation protocol of DA3~\cite{lin2025depth}, we evaluate on HiRoom, ETH3D~\cite{schops2017multi}, 7Scenes~\cite{shotton2013scene}, and ScanNet++~\cite{yeshwanth2023scannet++}. These evaluation sequences are held out from training at the scene/sequence level.
We compare with offline full-attention models
(VGGT~\cite{wang2025vggt}, MapAnything~\cite{keetha2025mapanything},
Pi3X~\cite{wang2025pi3}, DA3) and online streaming models
(CUT3R~\cite{wang2025continuous}, StreamVGGT~\cite{zhuo2025streaming},
Wint3R~\cite{li2025wint3r}, Stream3R~\cite{lan2025stream3r},
LingBot-Map~\cite{chen2026geometric}). We adopt DA3's evaluation
protocol and use TSDF fusion~\cite{curless1996volumetric} for 3D
consistency. We report AUC@3/AUC@30 for pose accuracy and F1-score for
3D reconstruction quality.


\textbf{Camera Pose Estimation.}
As shown in Tab.~\ref{tab:geometry_benchmark}(a), our method achieves the best average performance among streaming models, indicating that it can recover camera motion reliably from sequential inputs. While full-attention models such as DA3 still benefit from bidirectional reasoning over the entire sequence, our method substantially reduces the gap under a causal streaming setting and even outperforms several full-attention baselines.

\input{table/geom}
\begin{figure}[th]
    \vspace{-3mm}
    \centering
    \includegraphics[width=0.96\textwidth]{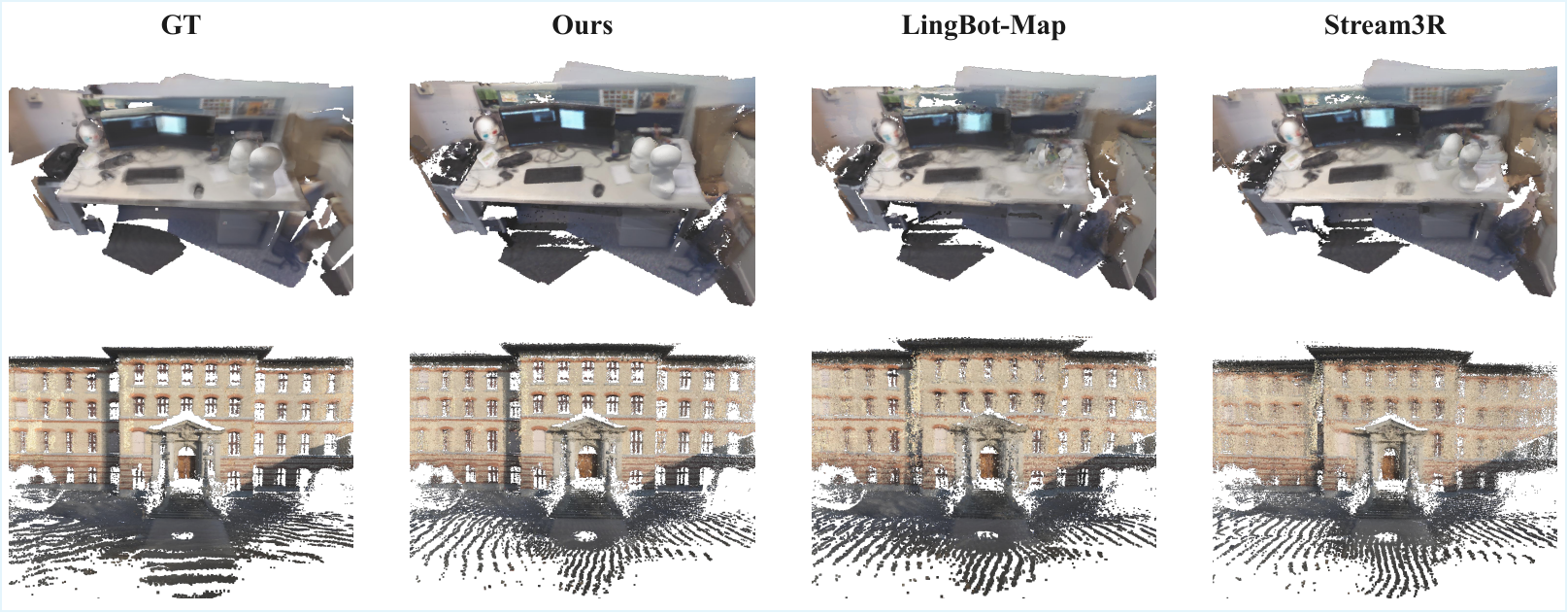}
    \caption{\textbf{Qualitative comparison of 3D reconstruction on the 7Scenes and ETH3D datasets.} }
    \label{fig:geo_compare}
    \vspace{-3mm}
\end{figure}

\textbf{3D Reconstruction.} Tab.~\ref{tab:geometry_benchmark}(b) reports 3D reconstruction F1-scores with predicted poses (w/o p.) and ground-truth poses (w/ p.). MapAnything, Pi3X, DA3, and our method accept ground-truth poses during inference, while the other methods use them only for evaluation-time fusion. In both settings, our method outperforms all streaming baselines, showing strong 3D consistency under sequential inference. With ground-truth poses, it further improves reconstruction quality, suggesting that the model can effectively exploit streaming camera inputs. Fig.~\ref{fig:geo_compare} shows qualitative comparisons with LingBot-Map and Stream3R using TSDF-fused point clouds, with invalid ground-truth depth regions masked out.


\subsection{Evaluation of Instance Spatial Tracking}
\label{sec:eval_tracking}

\begin{figure}[t]
    \centering
    \includegraphics[width=0.96\textwidth]{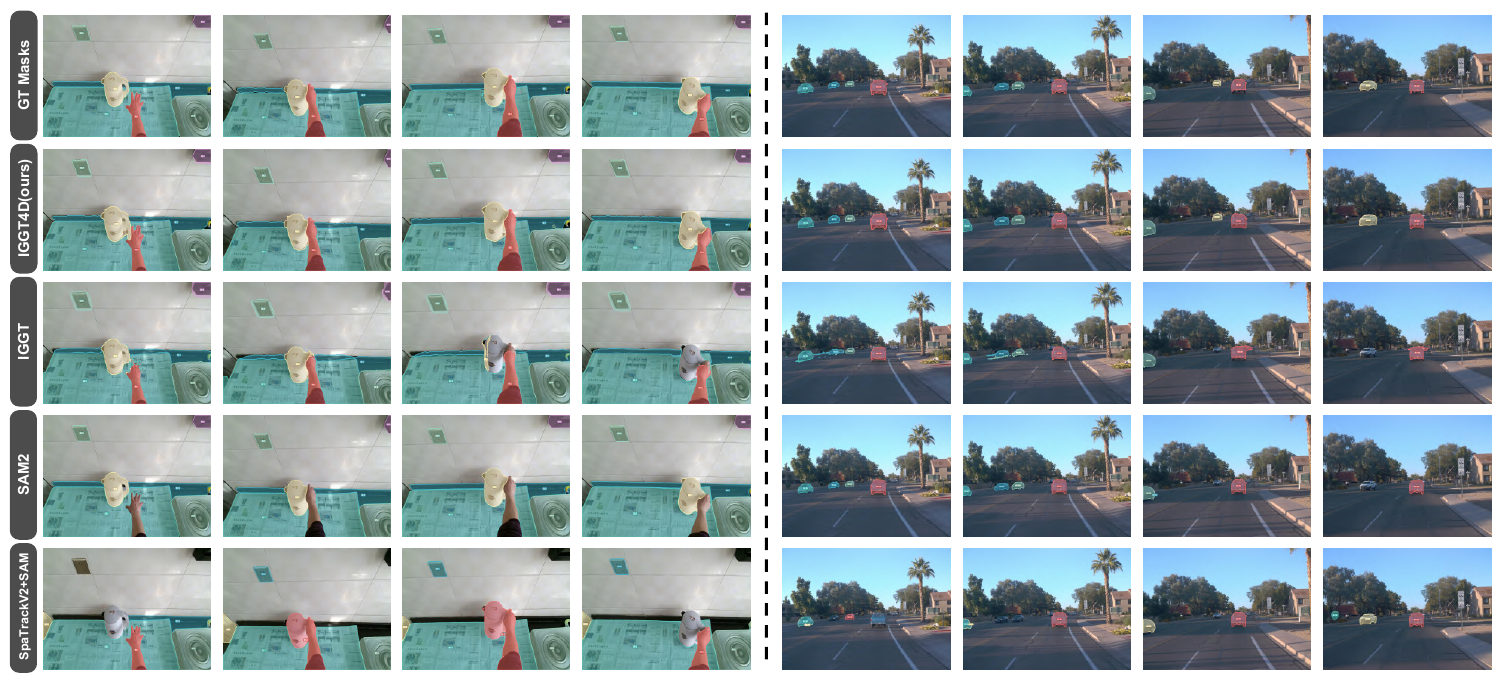}
    \caption{\textbf{Qualitative visualization results of Instance Spatial Tracking.}}
    \label{fig:ins_vis}
    \vspace{-0.2in}
    \vspace{-3mm}
\end{figure}

For instance spatial tracking and open-vocabulary semantic segmentation, all evaluation sequences are sampled from held-out splits that are disjoint from the training data at the scene or sequence level.
Our evaluation is conducted on three dynamic datasets (HOI4D~\cite{liu2022hoi4d}, Waymo~\cite{mei2022waymo}, and PointOdyssey~\cite{zheng2023pointodyssey}) and one static dataset (ScanNet++). Following the protocol in IGGT~\cite{li2025iggt}, we compare against SpaTrackerV2~\cite{xiao2025spatialtrackerv2}+SAM~\cite{kirillov2023segment}, SAM2~\cite{ravi2024sam}, and IGGT itself, and report Temporal mIoU (T-mIoU) and Temporal Success Rate (T-SR).
We evaluate the methods under both long- and short-sequence settings. For long sequences, all datasets consist of 100 frames, except for HOI4D, which uses 40 frames. As shown in Tab.~\ref{tab:instance_tracking}, IGGT suffers from out-of-memory (OOM) issues when processing long sequences. In contrast, benefiting from the proposed streaming inference and clustering mechanisms, our method can scale to 4D long sequences. We present the qualitative visualization results of Instance Spatial Tracking in Fig.~\ref{fig:ins_vis}.

\input{table/instance}

\subsection{Evaluation of Open-Vocabulary Semantic Segmentation}
\label{sec:eval_semantic}

\begin{figure}[tbp]
    \vspace{-1mm}
    \centering
    \includegraphics[width=0.96\textwidth]{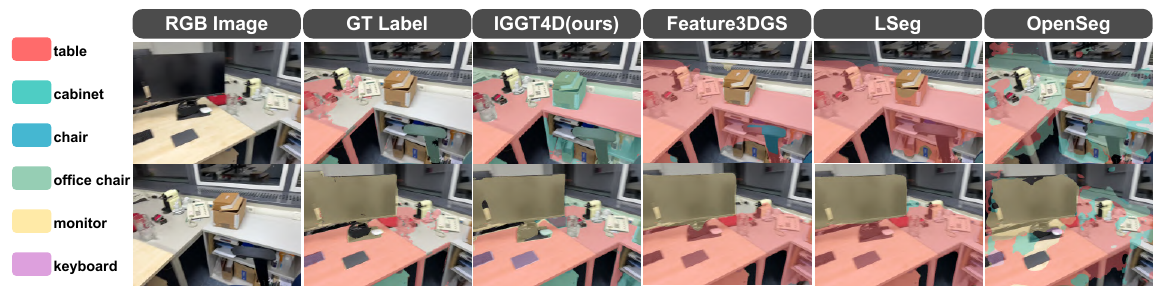}
    \caption{\textbf{Qualitative visualization results of Open-Vocabulary Semantic Segmentation.}}
    \label{fig:sem_vis}
    \vspace{-1mm}
\end{figure}

We evaluate on Waymo and ScanNet++. Baselines include 2D vision-language models (OpenSeg~\cite{ghiasi2022scaling}, LSeg~\cite{li2022language}) and 3D semantic reconstruction methods (per-scene Feature-3DGS~\cite{zhou2024feature} and feed-forward IGGT~\cite{li2025iggt}). As shown in Tab.~\ref{tab:sequence_comparison_oom}, our method achieves the highest mIoU and mAcc, demonstrating stronger semantic segmentation performance under both long- and short-sequence settings in both dynamic outdoor and complex static indoor scenes. Fig.~\ref{fig:sem_vis} shows qualitative visualization results.
\input{table/ov_sem}

\subsection{Ablation Study and Streaming Clustering Efficiency}
\label{sec:ablation}
We conduct the ablation study on ScanNet++ using a lightweight model, as shown in Tab.~\ref{tab:ablation_scannetpp}. Removing Geometry-Aware Attention (Geo-Attn) degrades instance and semantic performance, demonstrating that instance features benefit from geometric priors. Furthermore, without First-Frame Geometric Normalization (FF-Norm), severe scale ambiguity arises during streaming reconstruction. This disrupts the unified geometry-instance representation, leading to a decline across all metrics.

\textbf{Streaming Clustering Efficiency.} Tab.~\ref{tab:efficiency_comparison} compares the time and GPU memory consumption of our streaming clustering against the offline HDBSCAN ~\cite{mcinnes2017hdbscan} used in IGGT at a resolution of $504 \times 336$. Benefiting from a lightweight clustering codebook and frame-by-frame cosine clustering, our method maintains a constant memory footprint ($\sim$0.7 GB) and scales linearly in time.

\input{table/ablation_and_cluster}

%% file: table/geom.tex
\begin{table*}[!t]
  \vspace{-6mm}
  \centering
  \caption{\textbf{Geometry benchmark on HiRoom, ETH3D, 7Scenes, and ScanNet++.} (a) Camera pose estimation (AUC@3, AUC@30). (b) 3D reconstruction (F1-score) without (w/o p.) and with (w/ p.) ground-truth camera poses. Methods are grouped into \emph{full-attention} (offline) and \emph{streaming} (online). Best and second-best results are highlighted in \textbf{bold} and \underline{underline}, respectively.}
  \label{tab:geometry_benchmark}
  \resizebox{\textwidth}{!}{
  \begin{tabular}{l cccccccccc}
    \toprule
     Methods & \multicolumn{2}{c}{Avg} & \multicolumn{2}{c}{HiRoom} & \multicolumn{2}{c}{ETH3D} & \multicolumn{2}{c}{7Scenes} & \multicolumn{2}{c}{ScanNet++} \\
    \cmidrule(lr){2-3} \cmidrule(lr){4-5} \cmidrule(lr){6-7} \cmidrule(lr){8-9} \cmidrule(lr){10-11}
    \midrule
    \multicolumn{11}{l}{\textbf{\textit{(a) Camera Pose Estimation}}} \\
    \rowcolor{headerbg} & AUC@3$\uparrow$ & AUC@30$\uparrow$ & AUC@3$\uparrow$ & AUC@30$\uparrow$ & AUC@3$\uparrow$ & AUC@30$\uparrow$ & AUC@3$\uparrow$ & AUC@30$\uparrow$ & AUC@3$\uparrow$ & AUC@30$\uparrow$ \\
    \midrule
    \multicolumn{11}{l}{\textit{Full-attention (offline)}} \\
    VGGT        & \trd{0.4001} & \trd{0.8725} & \trd{0.4927} & \trd{0.8840} & \trd{0.2895} & 0.8166 & \trd{0.2293} & \trd{0.8465} & \trd{0.5890} & \trd{0.9429} \\
    MapAnything & 0.2436 & 0.8569 & 0.3069 & 0.8806 & 0.2347 & \trd{0.8250} & 0.1683 & 0.8256 & 0.2646 & 0.8966 \\
    Pi3X        & \snd{0.4683} & \snd{0.9096} & \snd{0.6662} & \snd{0.9476} & \snd{0.3453} & \snd{0.8749} & \snd{0.2526} & \snd{0.8607} & \snd{0.6090} & \snd{0.9552} \\
    DA3         & \best{0.6046} & \best{0.9352} & \best{0.8465} & \best{0.9735} & \best{0.4541} & \best{0.9184} & \best{0.2780} & \best{0.8671} & \best{0.8397} & \best{0.9818} \\
    \cmidrule(lr){1-11}
    \multicolumn{11}{l}{\textit{Streaming (online)}} \\
    CUT3R       & 0.1022 & 0.6977 & 0.1752 & 0.7438 & 0.1187 & 0.6611 & 0.0654 & 0.7177 & 0.0494 & 0.6683 \\
    StreamVGGT  & 0.1606 & 0.7579 & 0.1686 & 0.7330 & 0.1300 & 0.6866 & \trd{0.1354} & 0.8136 & \trd{0.2085} & 0.7985 \\
    Wint3R      & 0.1133 & 0.6937 & \trd{0.2522} & 0.7998 & 0.1219 & 0.6693 & 0.0251 & 0.6929 & 0.0541 & 0.6129 \\
    Stream3R    & \trd{0.1871} & \trd{0.7947} & 0.2510 & \trd{0.8183} & \trd{0.1621} & \trd{0.7143} & 0.1313 & \trd{0.8209} & 0.2041 & \trd{0.8252} \\
    LingBot-Map & \snd{0.3060} & \snd{0.8647} & \snd{0.3388} & \snd{0.8727} & \best{0.2796} & \best{0.8417} & \snd{0.1636} & \snd{0.8257} & \snd{0.4419} & \snd{0.9188} \\
    \rowcolor{oursb}Ours & \best{0.4464} & \best{0.8924} & \best{0.7223} & \best{0.9533} & \snd{0.2734} & \snd{0.8242} & \best{0.2435} & \best{0.8534} & \best{0.5467} & \best{0.9389} \\
    \midrule
    \multicolumn{11}{l}{\textbf{\textit{(b) 3D Reconstruction}}} \\
    \rowcolor{headerbg} & F1(w/o p.)$\uparrow$ & F1(w/ p.)$\uparrow$ & F1(w/o p.)$\uparrow$ & F1(w/ p.)$\uparrow$ & F1(w/o p.)$\uparrow$ & F1(w/ p.)$\uparrow$ & F1(w/o p.)$\uparrow$ & F1(w/ p.)$\uparrow$ & F1(w/o p.)$\uparrow$ & F1(w/ p.)$\uparrow$ \\
    \midrule
    \multicolumn{11}{l}{\textit{Full-attention (offline)}} \\
    VGGT        & \trd{0.5775} & \trd{0.6477} & \trd{0.5549} & \trd{0.6676} & \trd{0.6249} & 0.7198 & 0.4640 & \trd{0.5048} & \trd{0.6662} & 0.6985 \\
    MapAnything & 0.4526 & 0.6407 & 0.4726 & 0.5904 & 0.5722 & \trd{0.7596} & \trd{0.4647} & 0.5016 & 0.3007 & \trd{0.7112} \\
    Pi3X        & \snd{0.6740} & \snd{0.7756} & \snd{0.7780} & \snd{0.8969} & \snd{0.7014} & \snd{0.8331} & \snd{0.4742} & \best{0.5740} & \snd{0.7424} & \snd{0.7986} \\
    DA3         & \best{0.7430} & \best{0.7980} & \best{0.8819} & \best{0.9563} & \best{0.8098} & \best{0.8689} & \best{0.5005} & \snd{0.5653} & \best{0.7796} & \best{0.8015} \\
    \cmidrule(lr){1-11}
    \multicolumn{11}{l}{\textit{Streaming (online)}} \\
    CUT3R       & 0.3644 & 0.4060 & 0.2795 & 0.1993 & 0.5941 & 0.6467 & 0.3221 & 0.3711 & 0.2621 & 0.4071 \\
    StreamVGGT  & 0.2703 & 0.3377 & 0.1162 & 0.2179 & 0.3730 & 0.4275 & 0.3119 & 0.3567 & 0.2800 & 0.3486 \\
    Wint3R      & 0.4260 & 0.4550 & \snd{0.4853} & 0.2969 & \trd{0.6167} & \snd{0.7447} & 0.3090 & 0.3901 & 0.2929 & 0.3885 \\
    Stream3R    & \trd{0.4551} & \trd{0.5504} & 0.4359 & \snd{0.6435} & 0.5432 & 0.6400 & \snd{0.4426} & \snd{0.4739} & \trd{0.3986} & \trd{0.4443} \\
    LingBot-Map & \snd{0.5228} & \snd{0.6082} & \trd{0.4749} & \trd{0.5973} & \snd{0.6615} & \trd{0.7394} & \trd{0.3794} & \trd{0.4726} & \snd{0.5756} & \snd{0.6236} \\
    \rowcolor{oursb}Ours & \best{0.6678} & \best{0.7225} & \best{0.8015} & \best{0.8501} & \best{0.7186} & \best{0.7724} & \best{0.4975} & \best{0.5441} & \best{0.6535} & \best{0.7236} \\
    \bottomrule
  \end{tabular}
  }
  \vspace{-6mm}
\end{table*}

%% file: table/instance.tex
\begin{table*}[t]
  \vspace{-3mm}
  \centering
  \caption{\textbf{Instance spatial tracking on HOI4D, Waymo, ScanNet++, and PointOdyssey.} Evaluated under two settings: (a) long sequences and (b) short sequences. Best and second-best results are highlighted in \textbf{bold} and \underline{underline}, respectively.}
  \label{tab:instance_tracking}
  \resizebox{\textwidth}{!}{
  \begin{tabular}{l cc cc cc cc cc}
    \toprule
    Methods & \multicolumn{2}{c}{Avg} & \multicolumn{2}{c}{HOI4D} & \multicolumn{2}{c}{Waymo} & \multicolumn{2}{c}{ScanNet++} & \multicolumn{2}{c}{PointOdyssey} \\
    \cmidrule(lr){2-3} \cmidrule(lr){4-5} \cmidrule(lr){6-7} \cmidrule(lr){8-9} \cmidrule(lr){10-11}
    \midrule
    \multicolumn{11}{l}{\textbf{\textit{(a) Long sequences ($\sim$100 frames, HOI4D: 40 frames)}}} \\
    \rowcolor{headerbg} & T-mIoU$\uparrow$ & T-SR$\uparrow$ & T-mIoU$\uparrow$ & T-SR$\uparrow$ & T-mIoU$\uparrow$ & T-SR$\uparrow$ & T-mIoU$\uparrow$ & T-SR$\uparrow$ & T-mIoU$\uparrow$ & T-SR$\uparrow$ \\
    \midrule
    SpaTrackerV2+SAM & \trd{43.73} & \snd{59.28} & \trd{64.40} & \trd{74.89} & \trd{42.13} & \trd{53.04} & \snd{21.86} & \snd{39.49} & \trd{46.52} & \trd{69.68} \\
    SAM2             & \snd{45.38} & \trd{58.65} & \snd{65.41} & \snd{78.34} & \snd{47.62} & \snd{56.93} & \trd{12.88} & \trd{23.24} & \snd{55.62} & \snd{76.07} \\
    IGGT             & \oom & \oom & \oom & \oom & \oom & \oom & \oom & \oom & \oom & \oom \\
    \rowcolor{oursb}Ours & \best{58.84} & \best{85.41} & \best{78.44} & \best{95.60} & \best{59.35} & \best{88.04} & \best{33.19} & \best{61.35} & \best{64.36} & \best{96.65} \\
    \midrule
    \multicolumn{11}{l}{\textbf{\textit{(b) Short sequences ($\sim$16 frames)}}} \\
    \rowcolor{headerbg} & T-mIoU$\uparrow$ & T-SR$\uparrow$ & T-mIoU$\uparrow$ & T-SR$\uparrow$ & T-mIoU$\uparrow$ & T-SR$\uparrow$ & T-mIoU$\uparrow$ & T-SR$\uparrow$ & T-mIoU$\uparrow$ & T-SR$\uparrow$ \\
    \midrule
    SpaTrackerV2+SAM & 46.84 & 63.19 & 61.94 & 71.20 & 47.57 & 57.92 & \snd{33.73} & \trd{59.47} & 44.12 & 64.17 \\
    SAM2             & \snd{51.75} & \trd{66.03} & \snd{65.87} & \trd{78.11} & \snd{60.65} & \trd{71.37} & 21.25 & 35.08 & \trd{59.21} & \trd{79.57} \\
    IGGT             & \trd{51.68} & \snd{84.65} & \trd{63.02} & \snd{87.14} & \trd{49.59} & \snd{88.56} & \trd{32.97} & \best{68.07} & \snd{61.13} & \snd{94.83} \\
    \rowcolor{oursb}Ours & \best{59.25} & \best{86.28} & \best{78.27} & \best{95.42} & \best{61.08} & \best{89.34} & \best{34.18} & \snd{63.47} & \best{63.45} & \best{96.88} \\
    \bottomrule
  \end{tabular}
  }
  \vspace{-3mm}
\end{table*}

%% file: table/ov_sem.tex
\begin{table*}[ht]
  \vspace{-3mm}
  \centering
  \caption{\textbf{Open-vocabulary semantic segmentation on Waymo and ScanNet++.} Evaluated under two settings: (a) long sequences and (b) short sequences. Best and second-best results are highlighted in \textbf{bold} and \underline{underline}, respectively.}
  \label{tab:sequence_comparison_oom}
  \resizebox{\textwidth}{!}{
  \begin{tabular}{l cc cc cc cc cc cc}
    \toprule
    \multirow{3}{*}{Methods} & \multicolumn{6}{c}{(a) Long sequences ($\sim$100 frames)} & \multicolumn{6}{c}{(b) Short sequences ($\sim$16 frames)} \\
    \cmidrule(lr){2-7} \cmidrule(lr){8-13}
    & \multicolumn{2}{c}{Avg} & \multicolumn{2}{c}{Waymo} & \multicolumn{2}{c}{ScanNet++} & \multicolumn{2}{c}{Avg} & \multicolumn{2}{c}{Waymo} & \multicolumn{2}{c}{ScanNet++} \\
    \cmidrule(lr){2-3} \cmidrule(lr){4-5} \cmidrule(lr){6-7} \cmidrule(lr){8-9} \cmidrule(lr){10-11} \cmidrule(lr){12-13}
    & mIoU$\uparrow$ & mAcc$\uparrow$ & mIoU$\uparrow$ & mAcc$\uparrow$ & mIoU$\uparrow$ & mAcc$\uparrow$ & mIoU$\uparrow$ & mAcc$\uparrow$ & mIoU$\uparrow$ & mAcc$\uparrow$ & mIoU$\uparrow$ & mAcc$\uparrow$ \\
    \midrule
    OpenSeg      & 19.16 & 45.77 & 15.14 & 44.10 & 23.17 & \snd{47.43} & 19.41 & 46.80 & 15.03 & 43.87 & 23.78 & \trd{49.73} \\
    LSeg         & \snd{31.95} & \snd{48.23} & \snd{35.52} & \snd{56.55} & \snd{28.37} & \trd{39.90} & \snd{32.06} & \trd{50.54} & \snd{34.82} & \trd{57.04} & \trd{29.29} & 44.03 \\
    Feature-3DGS & \trd{31.94} & \trd{48.15} & \trd{35.51} & \trd{56.49} & \trd{28.37} & 39.81 & 26.47 & 44.32 & \trd{34.81} & 56.88 & 17.80 & 32.90 \\
    IGGT         & \oom & \oom & \oom & \oom & \oom & \oom & \trd{31.16} & \snd{61.90} & 29.44 & \snd{62.42} & \snd{32.88} & \snd{61.38} \\
    \rowcolor{oursb}Ours & \best{40.08} & \best{73.84} & \best{41.63} & \best{78.07} & \best{38.52} & \best{69.61} & \best{39.11} & \best{72.25} & \best{40.68} & \best{75.51} & \best{37.53} & \best{68.98} \\
    \bottomrule
  \end{tabular}
  }
  \vspace{-3mm}
\end{table*}

%% file: table/ablation_and_cluster.tex
\begin{table*}[htbp]
\centering
\begin{minipage}[b]{0.49\textwidth}
    \caption{\textbf{Ablation study on ScanNet++.} Geo-Attn: Geometry-Aware Attention. FF-Norm: First-Frame Geometric Normalization.}
    \label{tab:ablation_scannetpp}
\end{minipage}\hfill
\begin{minipage}[b]{0.49\textwidth}
    \caption{\textbf{Clustering efficiency comparison.} Ours: Streaming clustering. IGGT: HDBSCAN. $N$ denotes the number of frames.}
    \label{tab:efficiency_comparison}
\end{minipage}

\vspace{1mm}

\begin{minipage}[t]{0.49\textwidth}
    \centering
    \resizebox{\textwidth}{!}{
    \begin{tabular}{lcccccc}
    \toprule
    \multirow{2}{*}{Model} & \multicolumn{2}{c}{Geometry} & \multicolumn{2}{c}{Instance} & \multicolumn{2}{c}{Semantic} \\
    \cmidrule(lr){2-3} \cmidrule(lr){4-5} \cmidrule(lr){6-7}
     & AUC@3 $\uparrow$ & F1-score $\uparrow$ & mIoU $\uparrow$ & Match Rate $\uparrow$ & mIoU $\uparrow$ & mAcc $\uparrow$ \\
    \midrule
    Full model                & 0.2984 & 0.4790 & 0.3098 & 0.6271 & 0.3666 & 0.6853 \\
    w/o Geo-Attn        & 0.2902 & 0.4778 & 0.2970 & 0.6074 & 0.3559 & 0.6796 \\
    w/o FF-Norm & 0.2334 & 0.3542 & 0.2873 & 0.6019 & 0.3480 & 0.6768 \\
    \bottomrule
    \end{tabular}
    }
\end{minipage}\hfill
\begin{minipage}[t]{0.49\textwidth}
    \centering
    \resizebox{\textwidth}{!}{
    \begin{tabular}{lcccccccc}
    \toprule
    \multirow{2}{*}{Method} & \multicolumn{2}{c}{$N = 8$} & \multicolumn{2}{c}{$N = 16$} & \multicolumn{2}{c}{$N = 32$} & \multicolumn{2}{c}{$N = 100$} \\
    \cmidrule(lr){2-3} \cmidrule(lr){4-5} \cmidrule(lr){6-7} \cmidrule(lr){8-9}
     & Time(s) & Mem(GB) & Time(s) & Mem(GB) & Time(s) & Mem(GB) & Time(s) & Mem(GB) \\
    \midrule
    Ours & 0.78  & 0.7  & 1.60 & 0.7  & 2.98 & 0.7 & 7.43 & 0.7 \\
    IGGT & 105.8 & 13.2 & 418.3  & 25.85 & OOM  & OOM & OOM  & OOM \\
    \bottomrule
    \end{tabular}
    }
\end{minipage}
\end{table*}

%% file: sections/conclusion.tex
\section{Conclusion}
\label{sec:conclusion}



We presented IGGT4D, a streaming geometry-instance framework for online 4D scene understanding. IGGT4D jointly predicts camera motion, scene geometry, and temporally consistent instance features via causal spatial-temporal modeling and geometry-grounded instance prediction. We also introduced InsScene4D-147K, a large-scale dataset with geometry-consistent instance annotations across real/synthetic and static/dynamic scenes. Experiments on reconstruction, pose estimation, instance tracking, and open-vocabulary segmentation show that IGGT4D improves object-level consistency while preserving scalable streaming inference.

\paragraph{Limitations.}
InsScene4D-147K scales 4D instance supervision, but broader data coverage is needed for stronger robustness and generalization. IGGT4D remains mainly supervised, motivating future work on self-supervised learning and larger curated supervision. Our evaluations are also perception-centric; integrating embodied actions and interaction feedback is a key step toward embodied scene understanding.

%% file: sections/appendix.tex
\section{Technical Appendices and Supplementary Material}
\label{app:supp}

\begingroup
\setlength{\parskip}{0.35em plus 0.12em minus 0.08em}
\setlength{\abovedisplayskip}{5pt plus 2pt minus 2pt}
\setlength{\belowdisplayskip}{5pt plus 2pt minus 2pt}
\setlength{\abovedisplayshortskip}{0pt plus 2pt}
\setlength{\belowdisplayshortskip}{3pt plus 2pt minus 2pt}
\setlength{\abovecaptionskip}{3pt}
\setlength{\belowcaptionskip}{1pt}
\setlength{\intextsep}{6pt plus 2pt minus 2pt}
\setlength{\textfloatsep}{8pt plus 2pt minus 2pt}
\setlength{\floatsep}{8pt plus 2pt minus 2pt}
\raggedbottom
\makeatletter
\renewcommand{\section}{%
  \@startsection{section}{1}{\z@}%
                {-1.5ex \@plus -0.4ex \@minus -0.2ex}%
                { 1.0ex \@plus  0.25ex \@minus  0.15ex}%
                {\large\bf\raggedright}%
}
\renewcommand{\subsection}{%
  \@startsection{subsection}{2}{\z@}%
                {-1.2ex \@plus -0.4ex \@minus -0.2ex}%
                { 0.45ex \@plus  0.12ex}%
                {\normalsize\bf\raggedright}%
}
\makeatother

\subsection{Visualization of the Automated Geometry-Guided Annotation Pipeline}

To demonstrate the effectiveness of our proposed automated geometry-guided annotation pipeline, Fig.~\ref{fig:compare_vis} visualizes the generated annotations across the three diverse data sources. Furthermore, for the RealEstate10K (Re10K) scenes, we provide a qualitative comparison between our refined annotations and the prior vanilla annotations from InsScene-15K~\cite{li2025iggt}.

\begin{figure}[!htbp]
    \centering
    \includegraphics[width=0.96\textwidth]{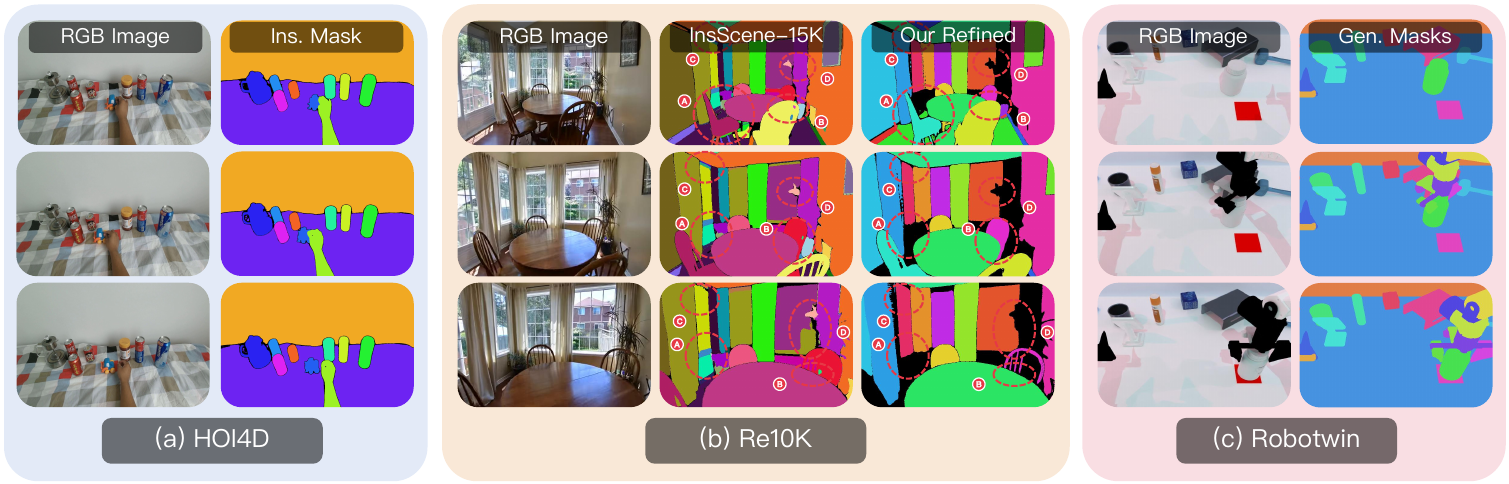}
    \caption{Qualitative evaluation of the automated geometry-guided annotation pipeline across three distinct sources (HOI4D, Re10K, and RoboTwin). For the Re10K scenes, the comparison illustrates how our approach overcomes the common failure modes observed in the vanilla InsScene-15K annotations: (a) over-segmentation, (b) false-positive segmentation, (c) identity switching, and (d) background leakage.}    \label{fig:compare_vis}
\end{figure}

\subsection{4D QA Scene Grounding}

To demonstrate the capability of IGGT4D in supporting complex spatial-temporal reasoning, we evaluate 4D QA scene grounding using Large Multimodal Models (LMMs)~\cite{yang2025qwen3, bai2025qwen3}. When provided with only RGB video streams, LMMs often struggle to track moving objects under challenging visual conditions. For example, as shown in Fig.~\ref{fig:vlm_demo}, when a picked-up bottle moves from left to right, its appearance closely blends with the background. Consequently, the LMMs erroneously infer that the bottle has been moved out of view based on its motion trajectory. However, when augmented with our 4D-consistent instance features or masks, the LMMs can robustly track the object, correctly determine that it remains within the field of view, and accurately segment the bottle.

\begin{figure}[!htbp]
    \vspace{-1mm}
    \centering
    \includegraphics[width=0.86\textwidth]{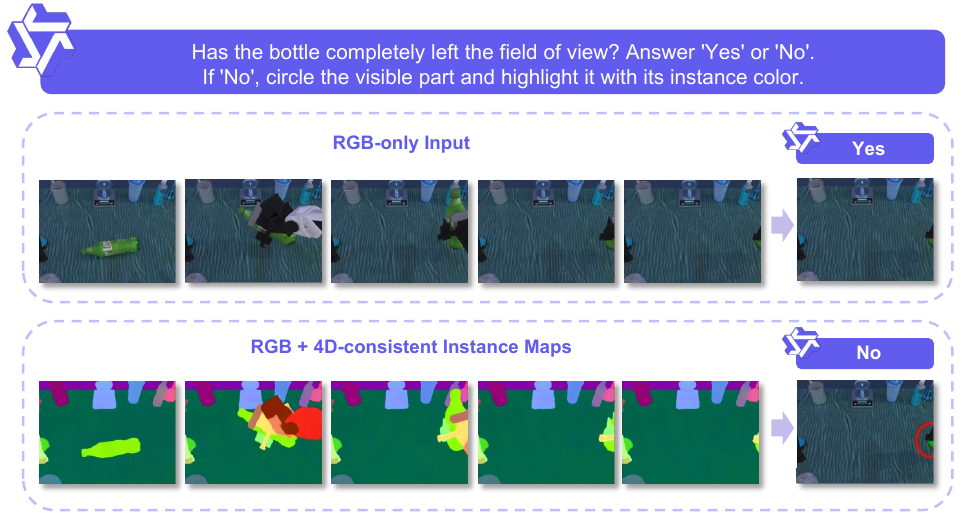}
    \caption{\textbf{4D QA Scene Grounding.} Comparison of spatial-temporal reasoning capabilities using LMMs. Given only RGB frames (top), the LMM fails to track the moving bottle as it blends into the background, erroneously concluding it has left the view. In contrast, by incorporating our 4D-consistent instance features (bottom), the LMM successfully tracks the object across frames, accurately grounds its location, and correctly answers the 4D semantic query despite challenging visual conditions.}
    \label{fig:vlm_demo}
    \vspace{-3mm}
\end{figure}

\FloatBarrier
\subsection{Explanation of First-Frame Geometric Normalization}

In our streaming training framework, we utilize the first-frame point cloud to normalize the geometric ground truth for the entire sequence. This design is crucial for preventing scale ambiguity. Consider an extreme scenario where global sequence-level normalization is applied instead. Suppose the model processes two different sequences that share an identical first frame. In the first sequence, the second frame captures a close-up of the ground, whereas in the second sequence, it captures an expansive open plaza. Under global normalization, the overall geometric scales of these two sequences would be drastically different. 

Consequently, when the streaming model processes the identical first frame in both cases, it is forced to regress entirely different scale values. From the model's perspective, observing only the first frame provides insufficient information to correctly infer the global scale of the yet-unseen future frames. This fundamentally leads to scale ambiguity and optimization conflicts. By anchoring the geometric normalization strictly to the first frame, we ensure that the scale is uniquely and consistently determined by the initial observation, completely eliminating this ambiguity and ensuring stable causal streaming training. Fig.~\ref{fig:ff_norm} compares streaming predictions \emph{without} versus \emph{with} first-frame geometric normalization (FF-Norm).

\begin{figure}[H]
    \centering
\includegraphics[width=0.98\textwidth]{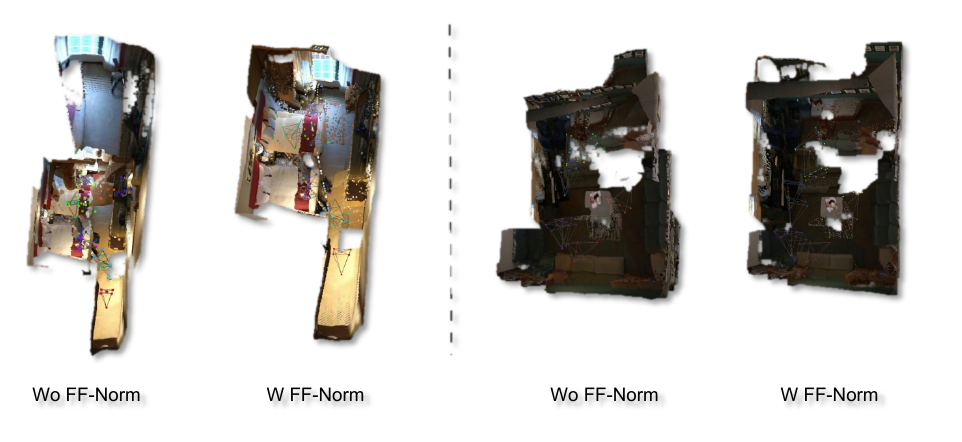}
    \caption{\textbf{First-frame geometric normalization (FF-Norm).} We compare streaming predictions \emph{without} versus \emph{with} first-frame geometric normalization.}
    \label{fig:ff_norm}
\end{figure}

\subsection{Visualization of Streaming Clustering Masks}

We further visualize instance-aware representations and the instance masks produced by our efficient streaming clustering. Fig.~\ref{fig:pca_streaming_masks} shows PCA projections of the predicted instance features together with the corresponding temporally consistent masks, illustrating stable identity tracking across frames without explicit motion modeling.

\begin{figure}[!htbp]
    \vspace{-3mm}
    \centering
    \includegraphics[width=0.98\textwidth]{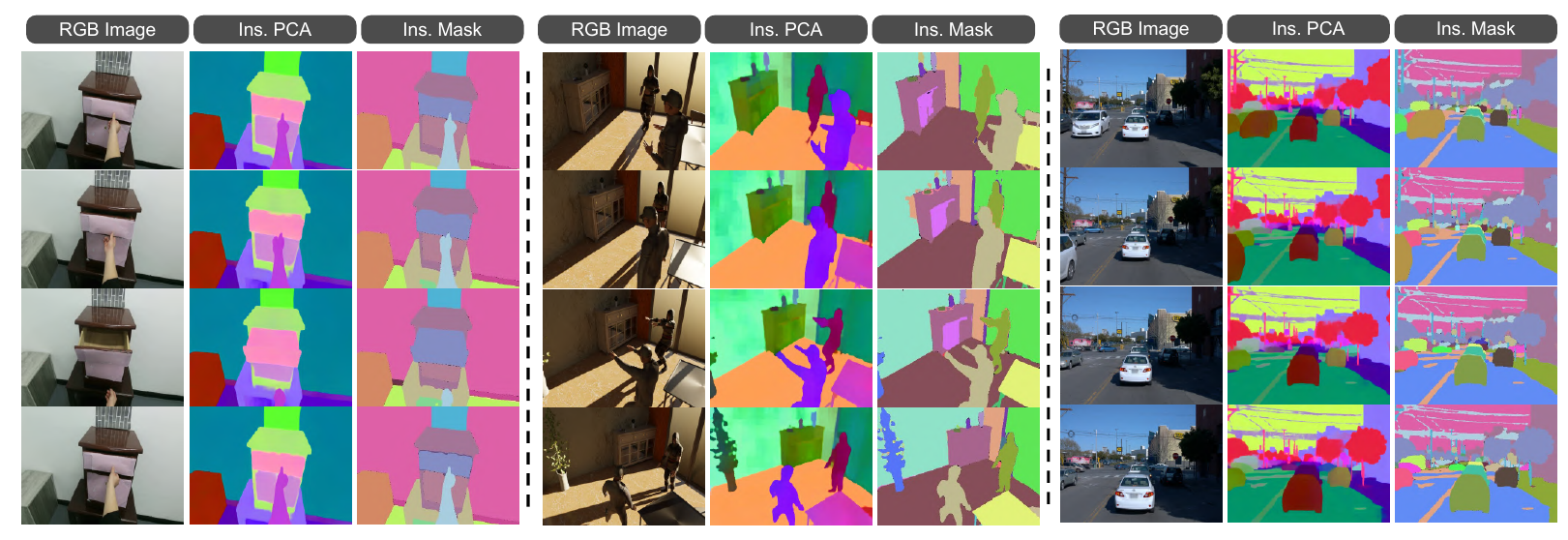}
    \caption{\textbf{Streaming clustering masks.} PCA visualization of instance features alongside masks obtained by our streaming clustering, demonstrating 4D-consistent instance segmentation in dynamic sequences.}
    \label{fig:pca_streaming_masks}
    \vspace{-3mm}
\end{figure}

\FloatBarrier
\subsection{Training Details}

Our model is initialized from the DA3-Giant~\cite{lin2025depth} architecture. We train IGGT4D on our constructed \textbf{InsScene4D-147K} dataset using 16 NVIDIA H20 GPUs. The training process is divided into two stages: the first 6 epochs focus exclusively on optimizing the streaming geometry, while the subsequent 9 epochs jointly optimize the geometry and instance branches. During training, the input image resolution varies between $504 \times 504$ and $504 \times 280$. To simulate streaming sequences of varying lengths, the total number of frames per iteration is fixed at 24, with the sequence length and batch size dynamically adjusted ($B \in \{12, 8, 6, 4, 3, 2\}$). The initial learning rate is set to $2 \times 10^{-4}$ for the instance branch and $1 \times 10^{-5}$ for the geometry. Both learning rates are decayed following a cosine annealing schedule. Additionally, we apply a 20\% probability of injecting ground-truth camera poses during the training. For all ablation studies, we adopt the DA3-Large architecture and train the models using 16 NVIDIA RTX 5090 GPUs.

\subsection{Experimental Details}

For \textit{instance spatial tracking}, we compare with SpaTrackerV2~\cite{xiao2025spatialtrackerv2}+SAM~\cite{kirillov2023segment}, SAM2~\cite{ravi2024sam}, and IGGT~\cite{li2025iggt}. Following the protocol in IGGT, we integrate SAM into SpaTrackerV2 by utilizing tracking points as prompts for dense segmentation, and modify SAM2 to support multi-view dense segmentation and tracking. We use HOI4D~\cite{liu2022hoi4d}, Waymo~\cite{mei2022waymo}, ScanNet++~\cite{yeshwanth2023scannet++}, and PointOdyssey~\cite{zheng2023pointodyssey}. HOI4D contains about 40 frames per scene and is manually annotated by us. Waymo and ScanNet++ each contain about 100 frames per scene; we use the official instance annotations and manually remove several small objects. For PointOdyssey, we use scenes with about 100 frames from the official test set, manually select a subset of masks, and merge them when needed.

For \textit{camera pose estimation and 3D reconstruction}, we compare our method with offline full-attention models (VGGT~\cite{wang2025vggt}, MapAnything~\cite{keetha2025mapanything}, Pi3X~\cite{wang2025pi3}, DA3~\cite{lin2025depth}) and online streaming models (CUT3R~\cite{wang2025continuous}, StreamVGGT~\cite{zhuo2025streaming}, Wint3R~\cite{li2025wint3r}, Stream3R~\cite{lan2025stream3r}, LingBot-Map~\cite{chen2026geometric}). Among them, MapAnything, Pi3X, DA3, and our method support the injection of ground-truth camera poses. Among the streaming methods, our method, CUT3R, StreamVGGT, and Stream3R perform frame-by-frame inference; Wint3R utilizes overlapping sliding windows with a window size of 4 and a stride of 2; and LingBot-Map initially applies full attention to the first 8 frames, subsequently reconstructing the remaining frames in a streaming manner.

Our evaluation datasets (HiRoom, ETH3D~\cite{schops2017multi}, 7Scenes~\cite{shotton2013scene}, and ScanNet++~\cite{yeshwanth2023scannet++}) and protocol follow the DA3 benchmark. To enable streaming inference, we reorder each sequence to ensure that every frame shares visual overlap with at least one preceding frame. We adopt the robust evaluation pipeline from DA3: predicted poses are aligned to the ground truth using RANSAC-based \texttt{evo} alignment, and the aligned poses alongside predicted depths are integrated via TSDF fusion to assess 3D consistency. For camera pose estimation, we report the Area Under the Curve (AUC) at error thresholds of $3^\circ$ (AUC@3) and $30^\circ$ (AUC@30), where accuracy is determined by the minimum of the Relative Rotation Accuracy (RRA) and Relative Translation Accuracy (RTA). For 3D reconstruction, we evaluate the Chamfer Distance (CD) and F1-score. Let $\mathcal{P}_{\mathrm{pred}}$ and $\mathcal{P}_{\mathrm{gt}}$ denote the predicted and ground-truth point clouds, respectively:
{\small
\[
  \begin{aligned}
    d(\mathbf{x}, \mathcal{S})
      &= \min_{\mathbf{y}\in\mathcal{S}} \|\mathbf{x} - \mathbf{y}\|_2, \\[2pt]
    \mathrm{Accuracy}
      &= \frac{1}{|\mathcal{P}_{\mathrm{pred}}|}
         \sum_{\mathbf{p}\in\mathcal{P}_{\mathrm{pred}}} d(\mathbf{p}, \mathcal{P}_{\mathrm{gt}}),
    &\quad
    \mathrm{Completeness}
      &= \frac{1}{|\mathcal{P}_{\mathrm{gt}}|}
         \sum_{\mathbf{g}\in\mathcal{P}_{\mathrm{gt}}} d(\mathbf{g}, \mathcal{P}_{\mathrm{pred}}), \\[2pt]
    \mathrm{Precision}
      &= \frac{1}{|\mathcal{P}_{\mathrm{pred}}|}
         \sum_{\mathbf{p}\in\mathcal{P}_{\mathrm{pred}}}
         \left[d(\mathbf{p},\mathcal{P}_{\mathrm{gt}})<\tau\right],
    &
    \mathrm{Recall}
      &= \frac{1}{|\mathcal{P}_{\mathrm{gt}}|}
         \sum_{\mathbf{g}\in\mathcal{P}_{\mathrm{gt}}}
         \left[d(\mathbf{g},\mathcal{P}_{\mathrm{pred}})<\tau\right], \\[2pt]
    \mathrm{CD}
      &= \frac{\mathrm{Accuracy} + \mathrm{Completeness}}{2},
    &
    \mathrm{F1\text{-}score}
      &= \frac{2\,\mathrm{Precision}\,\mathrm{Recall}}
              {\mathrm{Precision}+\mathrm{Recall}}.
  \end{aligned}
\]
}

We report the Chamfer Distance (CD) for 3D reconstruction in Tab.~\ref{tab:geometry_cd}. Our method still achieves the best average performance.

\input{table/geom_cd.tex}

\FloatBarrier
\subsection{Tri-DPT Geometry-Instance Head}
\begin{figure}[H]
    \centering
    \includegraphics[width=0.48\textwidth]{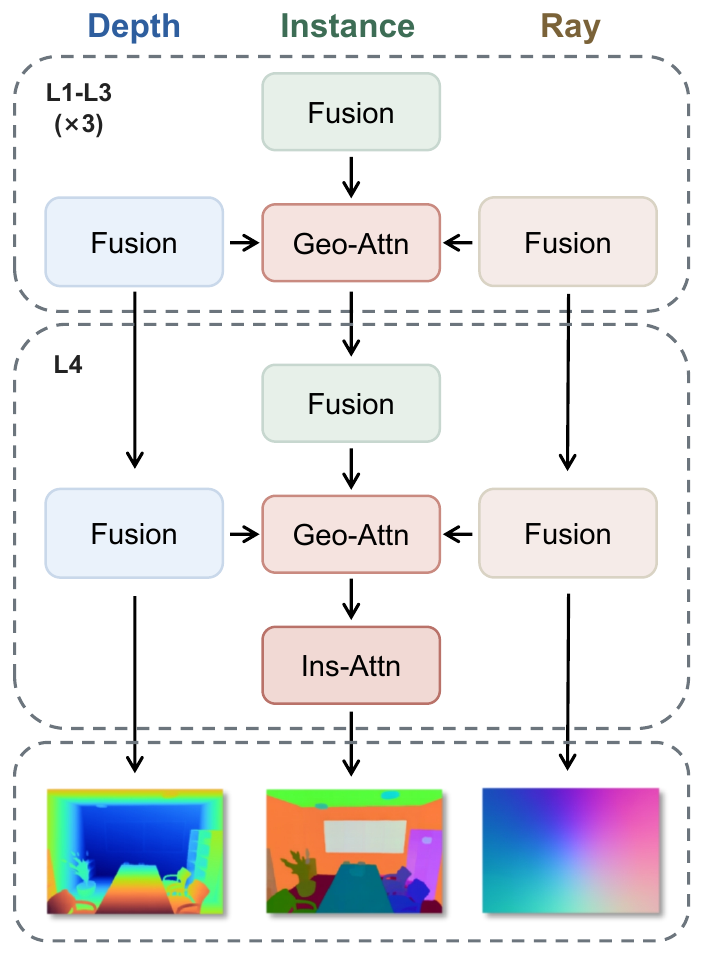}
    \caption{\textbf{Tri-DPT head.} The head jointly decodes depth, ray, and instance features, with geometry-aware attention injecting depth and ray priors into the instance branch.}
    \label{fig:tri_dpt_head}
\end{figure}

We design a Tri-DPT head to jointly decode geometry and instance representations from streaming multi-scale tokens $\{\mathbf{F}_t^{(l)}\}_{l=1}^{4}$. 
Rather than treating instance prediction as an independent add-on, Tri-DPT uses a unified DPT-style decoder with three coordinated output branches for the depth map $D_t$, ray map $R_t$, and instance feature map $S_t$. 
The multi-scale tokens are reassembled and fused progressively to recover spatial resolution, allowing dense geometric predictions and instance embeddings to be produced in the same spatially aligned representation.

As shown in Fig.~\ref{fig:tri_dpt_head}, instead of simply appending an independent instance head, the instance branch explicitly leverages multi-scale geometric features from the depth and ray branches. At the four fusion stages, depth features and ray features are fused as geometric priors and injected into the instance branch through geometry-aware attention. At the final stage, instance attention is further introduced to refine the instance features. As a result, the predicted instance features are not only driven by appearance cues, but are also tightly coupled with the underlying geometric structure. This design allows instance features to benefit from spatial-temporally consistent geometric constraints even under frame-by-frame streaming inputs, enabling stable 4D scene-level instance awareness.
\endgroup

%% file: table/geom_cd.tex
\begin{table}[!htbp]
  \centering
  \caption{\textbf{3D Reconstruction Chamfer Distance (CD) on HiRoom, ETH3D, 7Scenes, and ScanNet++.} Evaluated without (w/o p.) and with (w/ p.) ground-truth camera poses. Lower is better. Methods are grouped into \emph{full-attention} (offline) and \emph{streaming} (online). Best and second-best results are highlighted in \textbf{bold} and \underline{underline}, respectively.}
  \label{tab:geometry_cd}
  \resizebox{\textwidth}{!}{
  \begin{tabular}{l cccccccccc}
    \toprule
     Methods & \multicolumn{2}{c}{Avg} & \multicolumn{2}{c}{HiRoom} & \multicolumn{2}{c}{ETH3D} & \multicolumn{2}{c}{7Scenes} & \multicolumn{2}{c}{ScanNet++} \\
    \cmidrule(lr){2-3} \cmidrule(lr){4-5} \cmidrule(lr){6-7} \cmidrule(lr){8-9} \cmidrule(lr){10-11}
    \rowcolor{headerbg} & CD(w/o p.)$\downarrow$ & CD(w/ p.)$\downarrow$ & CD(w/o p.)$\downarrow$ & CD(w/ p.)$\downarrow$ & CD(w/o p.)$\downarrow$ & CD(w/ p.)$\downarrow$ & CD(w/o p.)$\downarrow$ & CD(w/ p.)$\downarrow$ & CD(w/o p.)$\downarrow$ & CD(w/ p.)$\downarrow$ \\
    \midrule
    \multicolumn{11}{l}{\textit{Full-attention (offline)}} \\
    VGGT        & \trd{0.2220} & \trd{0.1838} & 0.0968 & \trd{0.0646} & \trd{0.5669} & \trd{0.4664} & 0.1446 & 0.1269 & \trd{0.0798} & 0.0771 \\
    MapAnything & 0.2472 & 0.1883 & \trd{0.0863} & 0.0750 & 0.6478 & 0.4919 & \trd{0.1263} & \trd{0.1111} & 0.1283 & \trd{0.0754} \\
    Pi3X        & \snd{0.1889} & \snd{0.1458} & \snd{0.0449} & \snd{0.0307} & \snd{0.5326} & \snd{0.3990} & \best{0.1078} & \best{0.0897} & \snd{0.0702} & \best{0.0640} \\
    DA3         & \best{0.1688} & \best{0.1457} & \best{0.0355} & \best{0.0202} & \best{0.4499} & \best{0.3893} & \snd{0.1216} & \snd{0.1080} & \best{0.0680} & \snd{0.0651} \\
    \cmidrule(lr){1-11}
    \multicolumn{11}{l}{\textit{Streaming (online)}} \\
    CUT3R       & 0.4691 & 0.4235 & 0.7885 & 0.7858 & 0.7618 & 0.6409 & 0.1586 & 0.1454 & 0.1674 & 0.1219 \\
    StreamVGGT  & 0.3827 & 0.3410 & 0.2806 & 0.2036 & 0.9274 & 0.8623 & \trd{0.1573} & 0.1574 & 0.1655 & 0.1407 \\
    Wint3R      & \trd{0.3131} & 0.2629 & 0.2644 & 0.2410 & \trd{0.6525} & \snd{0.5368} & 0.1675 & \trd{0.1413} & 0.1680 & 0.1328 \\
    Stream3R    & 0.3223 & \trd{0.2569} & \trd{0.1278} & \trd{0.0911} & 0.7271 & \trd{0.5542} & 0.3046 & 0.2609 & \trd{0.1298} & \trd{0.1217} \\
    LingBot-Map & \snd{0.2431} & \snd{0.2158} & \snd{0.1019} & \snd{0.0796} & \snd{0.6451} & 0.5772 & \best{0.1352} & \snd{0.1203} & \snd{0.0904} & \snd{0.0860} \\
    \rowcolor{oursb}Ours & \best{0.1852} & \best{0.1556} & \best{0.0403} & \best{0.0354} & \best{0.4822} & \best{0.3976} & \snd{0.1370} & \best{0.1154} & \best{0.0814} & \best{0.0740} \\
    \bottomrule
  \end{tabular}
  }
\end{table}